\def\year{2021}\relax

\documentclass[letterpaper]{article} 
\usepackage[switch]{lineno}  %

\usepackage{aaai21}  
\usepackage{times}  
\usepackage{helvet} 
\usepackage{courier}  
\usepackage[hyphens]{url}  
\usepackage{graphicx} 
\usepackage{natbib}  
\usepackage{caption} 
\newcommand{\etal}{\textit{et al}.}
\newcommand{\ie}{\textit{i}.\textit{e}.}
\newcommand{\eg}{\textit{e}.\textit{g}.}
\newcommand{\tabincell}[2]{\begin{tabular}{@{}#1@{}}#2\end{tabular}}  
\usepackage{graphicx}
\usepackage{comment}
\usepackage{amsmath,amssymb} 
\usepackage{color}
\usepackage{amssymb}
\usepackage{algorithm}
\usepackage{algorithmic}
\usepackage{caption}
\usepackage{bm}
\usepackage{multirow}
\usepackage{subfigure}

\title{Learning Transferable Kinematic Dictionary for 
\\3D Human Pose and Shape Reconstruction}
\author {

        Ze Ma~\textsuperscript{\rm 1},
        Yifan Yao~\textsuperscript{\rm 1},
        Pan Ji\thanks{Work done while at NEC Labs America}~\textsuperscript{\rm 2}, 
        Chao Ma~\textsuperscript{\rm 1}\\
}
\affiliations {
    \textsuperscript{\rm 1} Shanghai Jiao Tong University \\
    \textsuperscript{\rm 2} OPPO US Research Center \\
    \{maze1234556, yao1010fan, chaoma\}@sjtu.edu.cn, peterji1990@gmail.com
}
\begin{document}

\maketitle

\begin{abstract}
Estimating 3D human pose and shape from a single image is highly under-constrained. To address this ambiguity, we propose a novel prior, namely kinematic dictionary, which explicitly regularizes the solution space of relative 3D rotations of human joints in the kinematic tree. Integrated with a statistical human model and a deep neural network, our method achieves end-to-end 3D reconstruction without the need of using any shape annotations during the training of neural networks. The kinematic dictionary bridges the gap between in-the-wild images and 3D datasets, and thus facilitates end-to-end training across all types of datasets. The proposed method achieves competitive results on large-scale datasets including Human3.6M, MPI-INF-3DHP, and LSP, while running in real-time given the human bounding boxes.
\end{abstract}

\section{Introduction}
3D human body reconstruction from monocular images has attracted increasing attention in recent years due to its wide range of real-world applications. A full representation of 3D human bodies usually consists of two parts: 3D pose (\eg, 3D keypoints/joints) and 3D body shape. Although great progress has been made over the years by incorporating human priors (\eg, joint angle limits~\cite{akhter2015pose,bogo2016keep}, bone length ratios~\cite{sun2017compositional,zhou2017towards} and body symmetry~\cite{moreno20173d}), the typical challenges still remain, in at least two aspects: (i) there lacks an end-to-end way to make full use of human priors for 3D {\it shape} reconstruction;
(ii) shape annotations are considerably more expensive than keypoint annotations. While shape annotations can be simulated by fitting a human model to located markers~\cite{loper2014mosh}, this is a non-trivial process and only applies to the MoCap dataset. To address these challenges, we aim to learn a kinematic dictionary, in which we distill the valuable human priors into compact knowledge, to facilitate 3D human pose and shape reconstruction under a semi-supervision manner, \ie, without using shape annotations during training the neural networks.

Existing 3D pose estimation approaches \cite{ramakrishna2012reconstructing,zhou2016deep,akhter2015pose} have made attempts to learn an over-complete dictionary of keypoint locations. With the learned pose dictionary, they recover 3D keypoints from ground-truth 2D keypoints using Orthogonal Matching Pursuit (OMP)~\cite{ramakrishna2012reconstructing}. Despite the demonstrated promising results, linearly combining the bases of the dictionary and sparse codes results in two limitations. First, the pose dictionary is learned over keypoint locations, which has high degrees of freedom and does not encode strict anthropometric priors. Second, it lacks constraints on optimizing sparse codes, leading to undesirable recovered poses that go beyond plausible space. To address these issues, we build a dictionary on the kinematic rotations rather than keypoint locations. To the best of our knowledge, we are the first to do so. Then we constrain the learning of sparse codes so that the recovered rotation always lies in the convex hull of the bases of the dictionary. This explicitly excludes the impossible rotations during inference. To achieve this, we propose a new objective function to learn the dictionary.

Equipped with the kinematic dictionary, we further embed them into the end-to-end deep learning framework. This is much different from the previous works~\cite{ramakrishna2012reconstructing,zhou2016deep,zhou20153d,wang2014robust,wangni2018monocular}, which emphasizes the optimization manner to use the dictionary. Our method focuses on how to benefit from the dictionaries together with a learning framework. Specifically, we decompose the kinematic rotations into the bases and predict the sparse codes to recover the kinematic rotations via deep neural networks. We avoid iterative optimization so that our method can run in real-time. 
To render human shape, we integrate the kinematic dictionary with a statistical human mesh model, SMPL~\cite{loper2015smpl}, which articulates the pose of the human by Forward Kinematics. As such, our proposed network directly outputs the human meshes. We also learn a shape dictionary to constrain the shape parameters of the SMPL model, so that the human can have the natural body type.
Compared to the recent end-to-end human mesh reconstruction work~\cite{omran2018neural,pavlakos2018learning,kanazawa2018end,kanazawa2019learning}, we do not use any shape annotations during the training of neural networks. 

In summary, the main contributions of this work are threefold: 
\begin{itemize}
    \item We propose to learn a novel kinematic dictionary to make full use of human priors. Our kinematic prior can be seamlessly integrated with other priors, such as a shape dictionary for SMPL, to achieve better results. 
    \item We integrate the kinematic dictionary with a statistical human mesh model to directly output 3D human shape without requiring shape annotations for training. Our method generalizes well to data with semi-supervision.
    \item We extensively evaluate our method on a variety of large-scale datasets, including Human3.6M~\cite{h36m_pami}, MPI-INF-3DHP~\cite{mehta2017monocular}, and UP-3D~\cite{lassner2017unite}. The proposed method achieves competitive results in comparison with the state of the arts and runs in real-time.
\end{itemize}

\section{Related Work}

In this section, we first discuss monocular 2D pose estimation and then review the topic of 3D pose estimation from a single image. At last, we discuss mesh-based 3D human reconstruction.

\paragraph{\bf Monocular 2D Pose Estimation.}
Estimating 2D pose from monocular images has achieved much success recently.
On one hand, the emergence of large-scale datasets, such as LSP~\cite{johnson2010clustered}, MPII~\cite{andriluka20142d}, MSCOCO~\cite{lin2014microsoft}, has greatly promoted the performance of the learning-based algorithms. With more data available, the major breakthrough emerges with deep convolutional neural networks~\cite{toshev2014deeppose,newell2016stacked,Cao_2017_CVPR,wei2016convolutional}. Motivated by the success in 2D pose estimation, early work on 3D pose estimation first estimates 2D keypoints and then lift them to 3D space~\cite{martinez2017simple,nie2017monocular}. However, the two-stage diagram is not robust to the noisy 2D estimation and cannot be optimized end-to-end. Our method is to embed a kinematic dictionary into a deep convolutional network. Without using any intermediate representation or any shape annotations, we achieve 3D human pose and body reconstruction in an end-to-end manner.
\paragraph{\bf Monocular 3D Pose Estimation.}
Early work formulates the task of 3D pose estimation from monocular images as an optimization problem equipped with the pose dictionary. Ramakrishna~\etal~\cite{ramakrishna2012reconstructing} learn an over-complete dictionary to model the distribution of 3D pose, then optimize sparse codes to fit the recovered 3D pose projection to the observed 2D keypoints.  This method motivates Zhou~\etal~\cite{zhou20153d,zhou2016sparseness} and Wang~\etal~\cite{wang2014robust} to develop better optimization algorithms.
Akhter~\etal~\cite{akhter2015pose} takes one step further to use conditioned joint angle priors in an optimization way to penalize the unreasonable poses. Despite the demonstrated success, optimization does not fully utilize the valuable pose priors and image clues. With the emergence of deep learning, considerable efforts have been made to deal with 3D pose estimation by learning deep regression networks~\cite{chen20173d,martinez2017simple,moreno20173d,nie2017monocular}. These methods typically estimate 2D keypoints first and then lift them to 3D keypoints. As mentioned above, these two-stage approaches are not robust to noisy 2D keypoints estimation and cannot be optimized end-to-end.  
So there comes recent work to directly estimate 3D pose from 2D images~\cite{rogez2016mocap,sun2017compositional,sun2018integral,luvizon20182d}. These works aim to bridge this gap to facilitate end-to-end training across 2D and 3D datasets. 
Among them, Zhou~\etal~\cite{zhou2016deep} manages to regress in the kinematic rotation space from monocular images and manually designs the rotation limits. On the contrary, we use a data-driven way to distill the kinematic knowledge from the MoCap dataset and then show they are also applicable in other datasets.

\paragraph{\bf 3D Human Reconstruction in Mesh.}
Human shape and pose reconstruction can be also viewed as an optimization problem. The pioneer methods \cite{bogo2016keep,lassner2017unite} use 2D keypoints, which are generated by off-the-shelf 2D pose estimators, to fit the SMPL model.
The learning-based methods directly regress SMPL parameters under different supervisions, such as vertices~\cite{kolotouros2019convolutional}, voxels~\cite{varol2018bodynet}, and 3D keypoints~\cite{kanazawa2018end}. On the other hand, the input representation, such as monocular RGB images~\cite{kanazawa2018end,tan2018indirect}, segmentations~\cite{omran2018neural}, as well as heatmaps and silhouettes~\cite{pavlakos2018learning} can be fed into the network. 
Recently, distinguished breakthrough has been made in this area to utilize Graph Convolution~\cite{kolotouros2019convolutional}, to synergize using dense annotation~\cite{Guler_2019_CVPR} or to integrate SMPLify with the regression model~\cite{kolotouros2019spin}.
Kolotouros~\etal~\cite{kolotouros2019convolutional} manages to directly regress vertices from single-view images. Guler~\etal~\cite{Guler_2019_CVPR} uses the dense annotations in DensePose~\cite{guler2018densepose} to synergize with the SMPL. Kolotouros~\etal~\cite{kolotouros2019spin} combines the regression model~\cite{kanazawa2018end} with SMPLify~\cite{bogo2016keep} to benefit from both the optimization manner and the learning manner. Dong~\etal~\cite{dong2020motion} asynchronous multi-view as a source to alleviate ill-pose issue.
Although these works have demonstrated promising results, they perform regression in the SMPL parameter space without any constraints. They thus need additional supervision on SMPL parameters or using the adversarial learning to regularize the solution space. And without any supervision on the SMPL parameters, it will be hard for networks to converge or to regress natural human meshes. We notice that only a very few datasets \cite{h36m_pami,akhter2015pose} provide such supervision signals, \ie, collecting data based on markers in the controlled environment. These datasets only have a single person and simple background contexts with a large domain gap from the real data. The recent 3D pose datasets~\cite{Saini_2019_ICCV,Joo_2015_ICCV,Joo_2017_TPAMI,mehta2017monocular} collect in-the-wild data without markers, resulting in semi supervisions. Hence it is of great importance to utilize the possible priors to constrain the kinematic solution space. Our learned dictionary well compresses the shape annotations from the MoCap data and can readily transfer to in-the-wild datasets. They greatly reduce the hidden solution of the kinematic rotation space and the shape space of SMPL as well. This not only accelerates the network convergence but also improves the results. 

\section{Technical Approach }
In this section, we first briefly review the SMPL model. Then we show the overview of our proposed model. After that, we carefully discuss how we learn the kinematic dictionary from data. We embed the learned dictionaries into the deep learning framework under semi-supervision. We use the dictionary to constrain the kinematic rotations in a plausible space. 

\subsection{Revisiting SMPL}

We use the SMPL model for its simplicity and flexibility. The SMPL model~\cite{loper2015smpl} provides a function \({\bf M} = M({\bm\Theta} , {\bm\beta})\), which takes as input the pose parameters \({\bm\Theta}\)  and shape parameters \(\bm\beta\), and returns the body mesh \({\bf M} \in \mathbb{R}^{3 \times P}\), with \(P = 6890\)  vertices. The pose \({\bm\Theta} \in \mathbb{R}^{4 \times K}\) is represented by the 3D relative rotations of \(K = 23\) joints in the human kinematic tree. The shape \({\bm\beta} \in \mathbb{R}^{10}\)  is modeled by the first 10 coefficients of PCA in a collected shape space. The SMPL function \(M({\bm\Theta} , {\bm\beta})\)  first adjusts the template shape conditioned on \({\bm\Theta}\)  and  \({\bm\beta}\),  and then articulates the joints via Forward Kinematics.  Note that this function is fully differentiable. It also provides a pre-defined matrix ${\bf Z}$ to get 3D keypoints from the mesh via linear transformation, \ie,
\begin{eqnarray}
{\bf X}_{3D} \in \mathbb{R}^{3 \times K} = {\bf Z} M({\bm\Theta} , {\bm\beta})\;. 
\end{eqnarray}

We assume a weak perspective camera model as the previous work~\cite{kanazawa2018end} so that our goal is to solve for a global rotation \({\bf R}\), a scaling factor \(s\) and a 2D translation vector \({\bf t}\), with which the 3D keypoints can be projected onto the image plane via
\begin{eqnarray}
{\bf X}_{2D} \in \mathbb{R}^{2 \times K} = s\Pi({\bf RX}_{3D}) + {\bf t}\;,
\label{camera}
\end{eqnarray}
where \(\Pi\) is an orthographic projection.

\begin{figure*}[t]
	\begin{center}
		\includegraphics[width=1\textwidth]{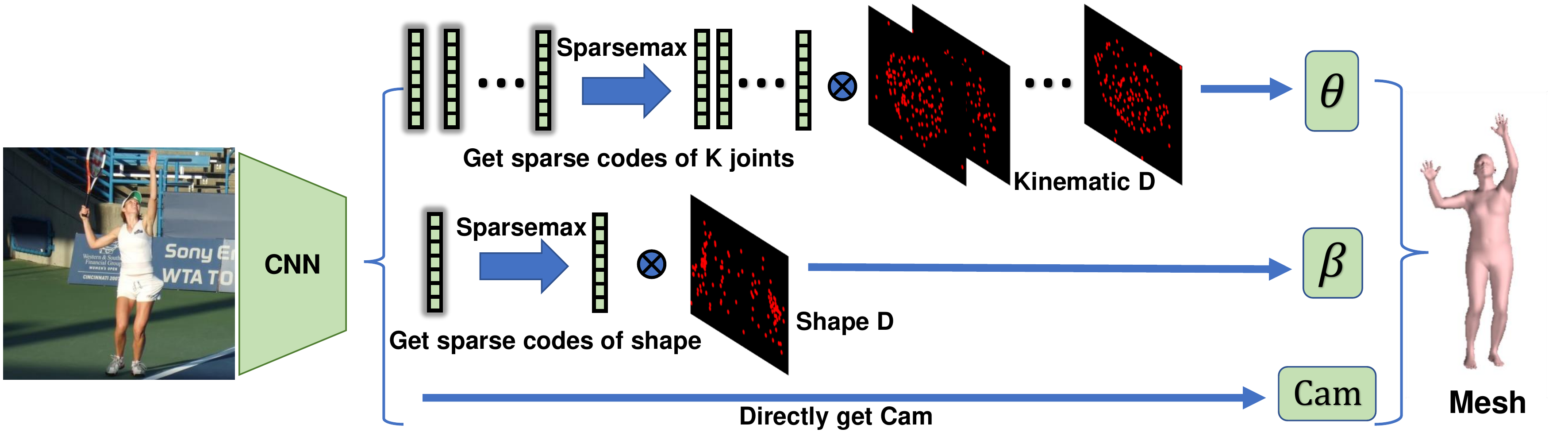}
	\end{center}
	\caption{\textbf{Overview of the proposed model.} The network predicts the sparse codes with the help of the sparsemax layer. Afterwards, it combines the sparse codes linearly with the pre-learned kinematic dictionary and the shape dictionary to recover the valid human joint kinematic rotations and shape parameters. With the directly regressed camera parameters, we finally use the SMPL model to reconstruct the 3D human mesh.}
	\label{Figure:Pipeline}
\end{figure*}

\subsection{Reconstructing Human Mesh via Kinematic Dictionary}

Our goal is to reconstruct the 3D human mesh from a single RGB image in the form of the SMPL model, which consists of a set of parameters $\{{\bm\Theta}, {\bm\beta}, {\bf R}, s, {\bf t}\}$. We resort to a data-driven method to solve this problem with the aid of a deep neural network and our proposed kinematic dictionary.

\paragraph{\bf Overall Idea.}
In the SMPL model, the pose \(\bm\Theta\) dominates human visualization. Therefore, regularizing the pose parameters is of utmost importance for human mesh reconstruction. To this end, we propose a simple kinematic dictionary prior to 
encode the space of all the plausible relative rotations of 3D human joints.
For each human joint, the model leverages a sparsemax layer~\cite{olshausen1997sparse} to predict the non-negative sparse codes \(\gamma\), whose $L_1$ norm equal to 1. 
The kinematic rotation \(\bm\theta \in \mathbb{R}^4\) of each joint in the quaternion representation is then recovered by a sparse linear combination of the bases in the kinematic dictionary. By doing so, \(\bm\theta\) is forced to lie in the convex hull of bases in the dictionary. We also extend our dictionary learning method readily to build a shape dictionary for \(\bm\beta\) to restrict the shape parameters of SMPL. We use the same camera setting as previous works~\cite{kanazawa2018end,kolotouros2019convolutional,omran2018neural} in Equation~\eqref{camera}. The only difference is that we use the representation proposed by Zhou~\etal~\cite{zhou2019continuity} for the global rotation, represented by the first two columns of the rotation matrix. 

In the following sections, we elaborate on the important modifications we make to sparse coding to adapt for learning kinematic rotation, 
and further extend it to the shape dictionary. We would like to point out that even though the shape dictionary is SMPL-dependent, the kinematic dictionary is model-agnostic and can be utilized in other shape models~\cite{SMPL-X:2019,anguelov2005scape,MANO:SIGGRAPHASIA:2017} that share a similar kinematic tree with SMPL. 

\begin{figure*}[t]
	\begin{center}
		\includegraphics[width=\textwidth]{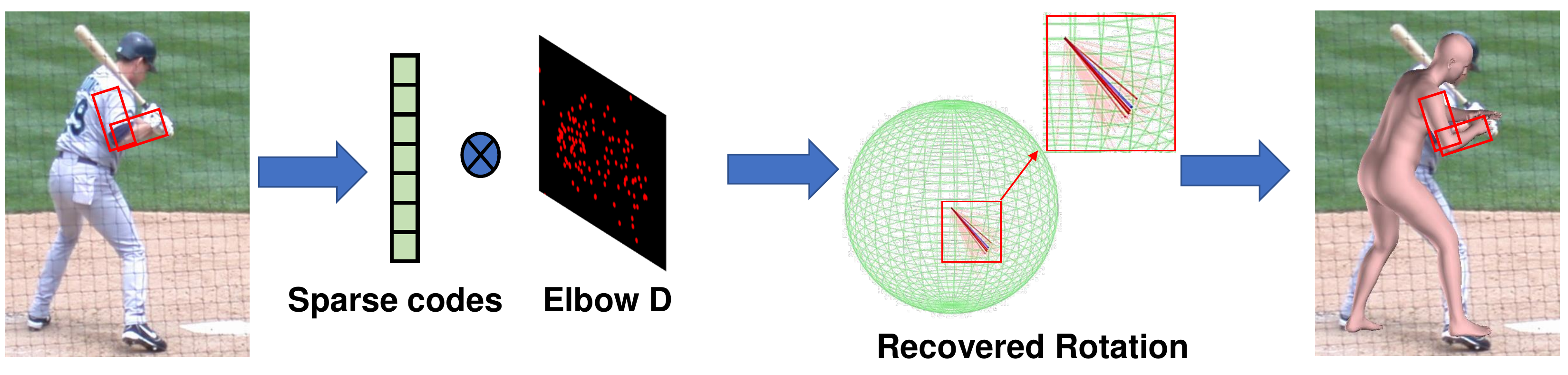}
	\end{center}
	\caption{Visualization of how to utilize the elbow joint dictionary to get the elbow rotation. The selected atoms in the dictionary are plotted in red and the inactive ones are in transparent red. The recovered rotation is in blue. We can see that, 1) the joint rotation is constrained to a small subspace of the whole rotation space, and 2) the selected atoms in the dictionary are close enough in favor of approximating Slerp by Lerp.}
	\label{Figure:vis_sparse_codes}
\end{figure*}

Figure~\ref{Figure:Pipeline} shows the overall pipeline of our proposed model, which can be trained in an end-to-end manner. In short, our network consists of a ResNet-50 backbone with a shared MLP to predict the pose codes, the shape codes, and the global camera parameters.

\paragraph{\bf Kinematic Dictionary.} As discussed above, in the heart of our method lies the construction of a kinematic dictionary which contains plausible 3D rotations. However, a dictionary of rotations also causes difficulties in pose reconstruction. It's well known that the 3D rotations lie on a nonlinear $SO(3)$ manifold, and a Linear Interpolation (Lerp) of them may lead to invalid composition~\cite{carfora2007interpolation}. As such, a Spherical Linear Interpolation (Slerp)~\cite{shoemake1985animating} is often demanded, which hinders the use of our kinematic dictionary in the end-to-end framework. Fortunately, we observe that the relative 3D rotations of human joints only span a small subspace of the complete $SO(3)$ manifold, which permits an approximate solution.

\paragraph{\bf From Slerp to Lerp via Sparsemax.}

Considering two valid rotation quaternions \({\bf q}_1, {\bf q}_2 \in \mathbb{R}^4\) in the dictionary, a new quaternion \({\bf q}\)  could be reconstructed by,
\begin{equation}
\begin{split}
{\bf q} \quad = \quad&Slerp(x_1, x_2; {\bf q}_1, {\bf q}_2) \\[1mm]
\quad    = \quad&\frac{\sin(x_1 \delta)}{\sin(\delta)} {\bf q}_1  + \frac{\sin(x_2\delta)}{\sin(\delta)}{\bf q}_2 \\[1mm]
\mathrm{s.t.} \quad  &x_1 + x_2 = 1\;, \quad x_1, x_2 \in [0, 1]\;, \
\end{split}
\end{equation}
where \(\delta\) is the angle subtended by the arc from \({\bf q}_1\) to \({\bf q}_2\).
In the limit as \(\delta \rightarrow 0\), which means \({\bf q}_1\)  and \({\bf q}_2\) are close enough, Slerp can be well approximated by Lerp, \ie, \({\bf q}  = x_1 {\bf q}_1  + x_2 {\bf q}_2\).
We further encourage a valid rotation recovery from Lerp by enforcing the dictionary codes to be sparse. We achieve this by the use of the sparsemax function~\cite{martins2016softmax}, 

\begin{equation}
\text { sparsemax }(\boldsymbol{z}):=\underset{\boldsymbol{p} \in \Delta^{K-1}}{\operatorname{argmin}}\|\boldsymbol{p}-\boldsymbol{z}\|^{2}
\end{equation}
where $\Delta^{K-1}:=\left\{\boldsymbol{p} \in \mathbb{R}^{K} | \mathbf{1}^{\top} \boldsymbol{p}=1, \boldsymbol{p} \geq \mathbf{0}\right\}$ is the $(K-1)$-dimensional simplex. This function returns the Euclidean projection of the input vector \boldsymbol{$z$}, which hits the boundary of the simplex, and then gets sparse output. With the distinctive feature of returning sparse posterior distributions, it can be used to identify which variables can contribute to the decision, making the model more interpretable.

	

We visualize the quaternions selected by sparse codes in Figure~\ref{Figure:vis_sparse_codes}, which empirically validates the effectiveness of the Slerp for our kinematic dictionary. It shows that the model learns to predict sparse codes that select closer quaternions to recover the target rotation. 


\paragraph{\bf Training Loss.}
During the training process, our loss function is, 
\begin{eqnarray}
\begin{split}
L &= L_{3D} + L_{2D}\;, \\
\end{split}
\label{eq:loss}
\end{eqnarray}
where each term is defined as
\begin{eqnarray}
\begin{split}
L_{3D} &= \| {\bf X}_{3D} - \hat {\bf X}_{3D}\|_2^2\;,\\
L_{2D} &= \| {\bf X}_{2D} - \hat {\bf X}_{2D}\|_2^2\;. 
\end{split}
\end{eqnarray}
Here we reuse the terms of ${\bf X}_{2D}$ and ${\bf X}_{3D}$ to represent 2D and 3D keypoints. We use $\hat {\bf X}_{3D}$ and $\hat {\bf X}_{2D}$, to refer to annotations. $L_{2D}$ captures the observation of 2D keypoints and $L_{3D}$ can help to recover depth ambiguity. Usually, these two losses are enough for the task of regressing 3D keypoint~\cite{pavlakos2017coarse,zhao2019semantic,chen20173d,martinez2017simple,moreno20173d,nie2017monocular}, but for shape recovery, models will fail without shape supervisions~\cite{kanazawa2018end}. This is because human shape has much more degrees of freedom than human keypoints, so by only constraining the keypoints to be accurate, there are still different ways to fit the human model, though most of the solutions will result in unnatural visualization.

To address this, previous methods~\cite{omran2018neural,pavlakos2018learning} introduce shape loss on the MoCap dataset where shape annotations are available. They put supervisions on SMPL parameters $\bm{\beta}$ and $\bm{\bm\Theta}$ or on Meshes ${\bf M}$. However, it becomes inapplicable when training on in-the-wild datasets because these annotations are no longer available. In our work, we avoid trainning neural networks with such supervision. Instead, we distill the kinematic and shape knowledge into our kinematic and shape dictionaries, and embed these dictionaries into the deep learning network as a sparse coding layer to constrain the solution space. We show that it not only works on the MoCap dataset, from which we distill the shape knowledge, but works on in-the-wild datasets as well, which demonstrates the knowledge encoded in the dictionaries is transferable.


\subsection{Learning Kinematic Prior via Online Batch Dictionary Learning}

We now show how to learn a kinematic dictionary using the prior from data.
We follow the online dictionary learning work~\cite{mairal2009online} from the sparse coding community for scalable learning, but make important changes to accommodate the fact that our kinematic dictionary contains rotations as its bases.
Specifically, we aim to solve an optimization problem as follows
\begin{equation}
\begin{split}
&\min\limits_{{\bf D}, \bf{w}}\frac{1}{2}\|\bm{\theta}-\bf{D}\bf{w}\|_2^2\;, \\ {\rm s.t.} \quad &\text{$\|{\bf w}\|_1 = 1$},\; {\bf D}= [{\bf d}_1\ {\bf d}_2\ \cdots\ {\bf d}_N],\\
& \|{\bf d}_i\|_2^2=1, \forall i = 1, \cdots, N\;.
\end{split}
\end{equation}
In~\cite{mairal2009online}, the sparsity of ${\bf w}$ is achieved by minimizing the $L_1$ norm of ${\bf w}$. However, since here we adopt the convex combination, which requires its  norm to be 1, so we propose to use the sparsemax function~\cite{martins2016softmax} in lieu of the $L_1$ norm, giving rise to a {\it smooth} objective, \ie, 
\begin{equation}
{L({\bf w},{\bf D})} = 
\frac{1}{2}\|\bm{\theta}-\mathbf{D}f(\mathbf{w})\|_2^2\;,
\end{equation}
where $f(\cdot)$ represents the smooth and convex sparsemax function.

To permit fast computation, we further propose the Online Batch Dictionary Learning (\textbf{OBDL}) that accepts data batches for data parallelism on GPUs. The learning objective becomes
\begin{equation}
\label{eq:obdl}
{L({\bf W},{\bf D})} = 
\frac{1}{2}\|\bm{\bar{\theta}}-\mathbf{D}f(\mathbf{W})\|_2^2\;,
\end{equation}
where \(\bm{\bar{\theta}} \in \mathbb{R}^{4\times b}\) contains a batch of quatations with the batch size \(b\), and \({\bf D} \in \mathbb{R}^{4 \times N}\) is the quaternion dictionary to be learned with \(N\) atoms. The sparse code \({\bm\Gamma} = f({\bf W}) \in \mathbb{R}^{N\times b} = [{\bm \gamma}_1\ {\bm \gamma}_2\ \cdots\ {\bm \gamma}_b]\) has the sparsemax normalized columns, each of which corresponds to one sample in the batch.

Following Mairal~\etal~\cite{mairal2009online}, we adopt an alternating strategy to minimize ${L({\bf W},{\bf D})}$, \ie, optimizing over one of the variables while fixing the other. We initialize the code \({\bf W}\) from the standard Gaussian distribution, which turns out to be a good initialization for the convergence. Note that \({\bf W}\) itself doesn't need to be sparse as the sparsity is guaranteed by the sparsemax \(f(\cdot)\). As in~\cite{mairal2009online}, we use matrices \({\bf A}\) and \({\bf B}\) to store the history information and \(\eta\) to control the momentum. In the step of optimizing the dictionary ${\bf D}$, we normalize its columns to keep the bases valid. Overall, the optimization problem for minimizing ${L({\bf W},{\bf D})}$ is non-convex; however, we empirically observe that the proposed alternating method converges fast with good local solutions. We summarize our dictionary learning method in Algorithm~\ref{alg1} and Algorithm~\ref{alg2}. 






\begin{algorithm}[t]
\caption{Learning Kinematic Prior via OBDL}
\label{alg1}
\scalebox{1}{
\begin{minipage}{\linewidth}

\begin{algorithmic}[1]
\REQUIRE $\bm{\bar{\theta}} \in \mathbb{R}^{4\times b}$, time step $T$, momentum $\eta \in [0,1)$.
\STATE Randomly initialize $ \mathbf{D}_0 \in \mathbb{R}^{ 4 \times N}$  from Gaussian.\\
Reset the history information: $\mathbf{A}_0 = {\bf I}_N , \mathbf{B}_0 = \mathbf{D}_0 $. 
\FOR{t = 1 to T}
\STATE Draw $\bm{\bar{\theta}}$ from the training data, randomly initialize $\mathbf{W} \in \mathbb{R}^{N \times b}$ from Gaussian.
\STATE Update $\mathbf{W}$ via gradient descent to minimize the objective function in Eq.~\eqref{eq:obdl}.

\STATE $\mathbf{A_t} = \eta \mathbf{A_{t-1}} + (1-\eta) \sum\limits_{k=1}^b{\bm \gamma}_k{\bm \gamma}_k^T$. 
\STATE $\mathbf{B_t} = \eta \mathbf{B_{t-1}} + (1-\eta) \sum\limits_{k=1}^b\bm{\bar{\theta}_t}[:,k] {\bm \gamma}_k^T$.
\STATE Update $\mathbf{D}_t$ using Algorithm~\ref{alg2} via block decent, with $\mathbf{D}_{t-1}$ as warm restart, to minimize Eq.~\eqref{eq:obdl}.

\ENDFOR
\STATE \textbf{Return} $\mathbf{D}_T$ (learned kinematic dictionary)
\end{algorithmic}
\end{minipage}}
\end{algorithm}

\begin{algorithm}[t]
\scalebox{1}{
\begin{minipage}{\linewidth}

\caption{Quaternion Dictionary Update}
\label{alg2}
\begin{algorithmic}[1]
\REQUIRE $\mathbf{D}_{t-1}$ (input dictionary), $\mathbf{A}_t$, $ \mathbf{B}_t $
\STATE  $\mathbf{D}_t \leftarrow \mathbf{D}_{t-1} $ (initialize with a warm restart)
\FOR{ $j = 1$ to $N$}
\STATE Update the $j$-th column to optimize Eq.~\eqref{eq:obdl}
$$
\mathbf{u}_j \gets \frac{1}{\mathbf{A}[j,j]}(\mathbf{b}_j - \mathbf{D}_{t}\mathbf{a}_j) + \mathbf{d}_j\;.
$$
\STATE Normalize $\mathbf{u}_j$ to get a valid unit quaternion $\mathbf{d}_j$,
$$
\mathbf{d}_j \gets \frac{1}{\|\mathbf{u}_j\|_2}\mathbf{u}_j\;,\; \mathbf{D}_{t}[:,j] = \mathbf{d}_j\;.
$$
\ENDFOR
\STATE \textbf{Return} $\mathbf{D}_t$ (updated dictionary)

\end{algorithmic}
\end{minipage}}

\end{algorithm}

\section{Experiments}
This section presents the experimental validation of our method. We first overview the implementation details. Then we report the overall performance on large-scale datasets as well as the qualitative results. Last, we present ablation studies.

\subsection{Implementation Details}
For dictionary learning, we learn a kinematic dictionary for each joint of the kinematic tree using in the SMPL model, resulting in a total of 23 dictionaries. We set the dictionary size to 128, and the batch size to 512. We also learn a shape dictionary for shape parameters for SMPL using MoCap data in the same way. 

To construct the neural network, we use Resnet-50~\cite{he2016deep} as the encoder as in~\cite{kanazawa2018end,omran2018neural}. For each input image, we take the 2048-D convolutional features before GAP (Global Average Pooling), then pass them into an MLP consisting of two 4096-neuron fully-connected layers and a leaky-relu layer. The MLP outputs the camera parameters and the initial codes of \(\bm{\theta}\)  and \(\bm{\beta}\). We feed the initial codes of  \(\bm{\theta}\)  and \(\bm{\beta}\) to the sparsemax layer in the neural network to obtain sparse codes. With corresponding dictionaries, we recover \(\bm{\theta}\)  and \(\bm{\beta}\) from their sparse codes and use SMPL to render the human mesh.

During training,  we use the 3D datasets of Human3.6M~\cite{h36m_pami}, MPI-INF-3DHP~\cite{mehta2017monocular}, and UP-3D~\cite{lassner2017unite} and in-the-wild 2D datasets of LSP, LSP-extended~\cite{johnson2010clustered}, MPII~\cite{andriluka20142d} and COCO~\cite{lin2014microsoft}, as in work~\cite{kanazawa2018end}. We follow previous work to train a generic model using all training data, and evaluate on each dataset respectively. Note that although we use the MoCap dataset of Human3.6M, we do not use any of its shape annotations, which can be seen from the loss function. 

More details are included in the supplementary material.



%


\subsection{Overall Performance}
%
Since our proposed dictionary learning scheme does not explicitly model human mesh with additional expert knowledge, \eg, post-processing~\cite{kolotouros2019spin}, temporal clues~\cite{kanazawa2019learning} or dense keypoints~\cite{Guler_2019_CVPR}, we fairly compare with the state-of-the-art methods built directly upon the SMPL model. To evaluate the transferability of the proposed dictionary, we report the scores on a variety of datasets that do not have shape annotations. However, since most of the previous methods report results in the large MoCap dataset Human3.6M, for a fair comparison,  we incorporate Human3.6M into training data but use only keypoint annotations, excluding the shape annotations provided by Mosh~\cite{loper2014mosh}. 


\paragraph{\bf Human3.6M.}
This large indoor MoCap dataset includes subjects performing different activities such as discussing, walking, and eating. By ruling out the shape annotations provided by Mosh, we only use keypoint annotations to supervise our model. We use subjects S1, S5, S6, S7, and S8 for training, and we evaluate on S9 and S11 following the standard protocols, P1 and P2. P1 uses S9 and S11 with all camera views as the test set and reports mean per joint position error (MPJPE) without Procrustes Alignment. P2 uses only frontal camera views of S9 and S11 and reports reconstruction error after Procrustes Alignment. 
Table~\ref{tab:h36m} shows that our method performs well against the state-of-the-arts.

\begin{table}[t]
\begin{center}
\scalebox{0.7}{
\setlength\tabcolsep{2mm}
\begin{tabular}{|l|c|c|}
\hline
Method &   P2$\downarrow$  &  P1$\downarrow$ \\
\hline
Akhter \etal \cite{akhter2015pose} &  181.1 &-\\
Ramakrishna \etal \cite{ramakrishna2012reconstructing} & 157.3 &-\\
Zhou \etal \cite{zhou20153d} &  106.7 &-\\
\hline
SMPLify \cite{bogo2016keep}&  82.3 &-\\
SMPLify from 91 kps\cite{lassner2017unite}&  80.7 &-\\
Pavlakos \etal \cite{pavlakos2018learning} &  75.9 &-\\
NBF  \cite{omran2018neural} &  59.9 &-\\
HMR \cite{kanazawa2018end} & 56.8 & 88.0\\
Kolotouros \etal  \cite{kolotouros2019convolutional} &  50.1 & 74.7\\
Ours(Semi) & \textbf{48.6} &  \textbf{66.6}\\
\hline
\end{tabular}}
\end{center}
\caption{Human3.6M results. The top 3 methods utilize a pose dictionary in the optimization manner and the others directly regress to SMPL parameters. We achieve competitive results in comparison with the state of the arts.}
\label{tab:h36m}
\end{table}
 

\begin{table}[t]

\begin{center}
\scalebox{0.6}{
\setlength\tabcolsep{2mm}
\begin{tabular}{|l|c c c|c c c|}
\hline
   & \multicolumn{3}{c|}{ After Rigid Alignment }& \multicolumn{3}{c|}{  Absolute } \\
\hline
Method & PCK$\uparrow$ & AUC$\uparrow$ & MPJPE$\downarrow$ & PCK$\uparrow$ & AUC$\uparrow$ & MPJPE$\downarrow$ \\
\hline
Mehta \etal \cite{mehta2017monocular} & - & - & - & 75.7 & 39.9 & \textbf{117.6} \\
VNect \cite{mehta2017vnect}& 83.9 & 47.3 & 98.0 &  \textbf{76.6} & \textbf{40.4} & 124.7 \\
\hline
HMR \cite{kanazawa2018end}& 86.3	& 47.8 & 89.8 & 72.9 & 36.5 & 124.2\\
Ours(Semi) & \textbf{91.7} & \textbf{53.1} & \textbf{75.2} & 74.2& 35.2 & 119.5\\
\hline
\end{tabular}}
\end{center}
\caption{MPI results. Our method performs better than HMR which utilizes GAN by a large margin after rigid alignment, and is comparable without alignment. 
}
\label{tab:MPI-INF-3DHP}
\end{table}

\paragraph{\bf MPI-INF-3DHP.}
This dataset is collected in a markerless system with a multi-view setup. 
Table~\ref{tab:MPI-INF-3DHP} shows that our method performs the best after alignment because the kinematic dictionary can well constrain human keypoints. Before rigid alignment, our method is comparable with HMR \cite{kanazawa2018end} because we use the same camera setting, which yields large global rotation and translation inconsistency.

\paragraph{\bf UP-3D.}
This dataset \cite{bogo2016keep} contains in-the-wild 2D images and also uses fitted SMPL parameters for evaluation. We report the results on the test set in Table~\ref{tab:UP-3D}. 


\paragraph{\bf LSP.}

We evaluate our method on the human body segmentation task. We do not use any segmentation labels in training. Table~\ref{tab:LSP} shows the segmentation accuracy and average F1 scores both on parts and binary segmentation. Our method is on par with SMPLify, 
which uses ground truth keypoints and segmentation as supervision. Our method performs better than HMR in part segmentation because of the kinematic and shape dictionary are more stable than training with adversarial learning. 
Our method runs in real-time, much faster than other methods.

\begin{table}[t]
\begin{center}
\setlength\tabcolsep{2mm}
\begin{tabular}{|l|c|}
\hline
Method  & Surface Error \\
\hline
Lassner \etal \cite{lassner2017unite} & 169.8 \\
Pavlakos \etal \cite{pavlakos2018learning}& 117.7 \\
\hline
Ours (semi-supervised) & \textbf{110.1}\\
\hline
\end{tabular}
\end{center}

\caption{UP-3D Results. UP-3D evaluates the shape error as mesh. Our favorable results show that the proposed dictionaries well distill the shape knowledge to transfer to the in-the-wild datasets. }
\label{tab:UP-3D}
\vspace{-4mm}
\end{table}
\begin{table}[t]
\begin{center}
\scalebox{0.7}{
\setlength\tabcolsep{2mm}
\begin{tabular}{|l|c c|c c|c|}
\hline
\multirow{2}{*}{Method} & \multicolumn{2}{c|}{Parts}& \multicolumn{2}{c|}{Fg vs Bg}&\multirow{2}{*}{Run Time}\\
 & Acc & F1 & Acc & F1 & \\
\hline
SMPLify from 91 kps \cite{bogo2016keep}&  88.82 & 0.67 & 92.17 & 0.88 & - \\
SMPLify \cite{lassner2017unite}&  87.71 & 0.64 & 91.89 & 0.88 & \(\sim\) 1 min\\
Decision Forests \cite{lassner2017unite} & 82.32 & 0.51 & 86.60 & 0.80& 0.13 sec\\
\hline
HMR \cite{kanazawa2018end} & 87.12	& 0.60 & 91.67 & 0.87 & 0.04 sec\\
Ours (semi-supervised) & 88.69 & 0.66 & 91.36 & 0.86 & \textbf{0.003 sec}\\
\hline
\end{tabular} }
\end{center}
\caption{LSP Results. Our method performs better than HMR, and comparably with optimization methods which use segmentation labels. 
}
\label{tab:LSP}
\vspace{-4mm}
\end{table}

\subsection{Qualitative Results}

\begin{figure*}[t]
\centering
\setlength\tabcolsep{1.5pt}
\begin{tabular}{ccc}
  	\includegraphics[width=0.32\textwidth]{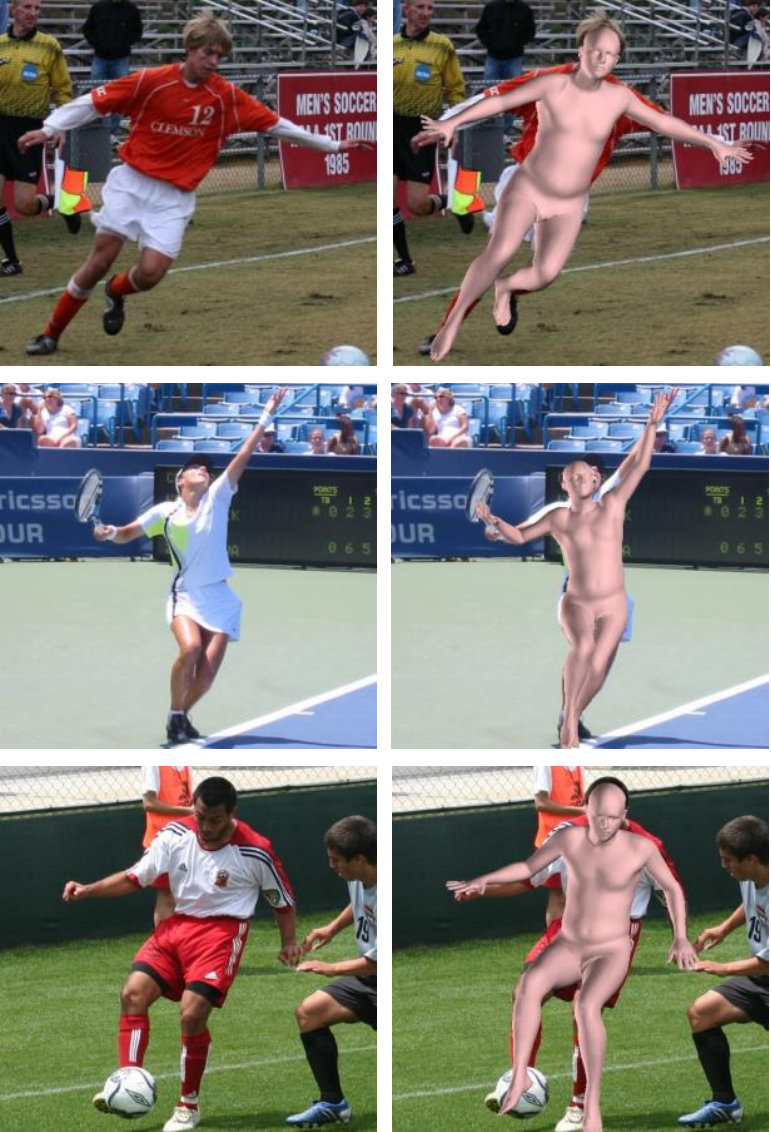} &
 	\includegraphics[width=0.32\textwidth]{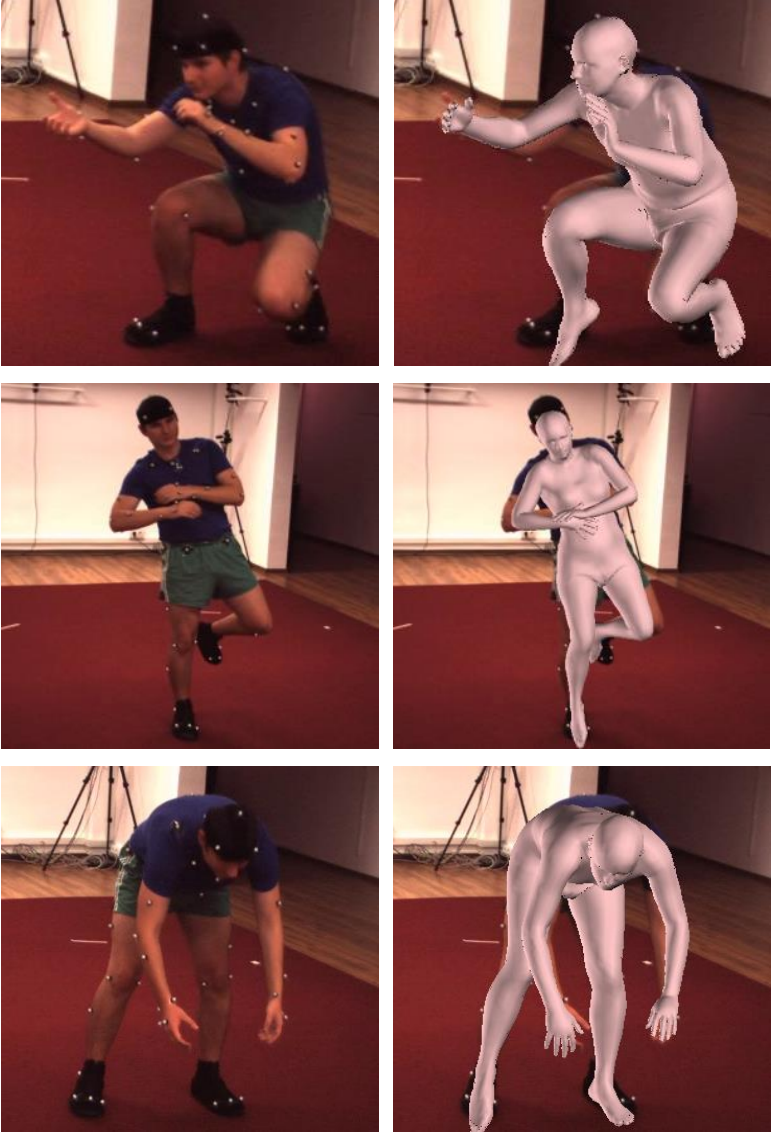} &
   	\includegraphics[width=0.32\textwidth]{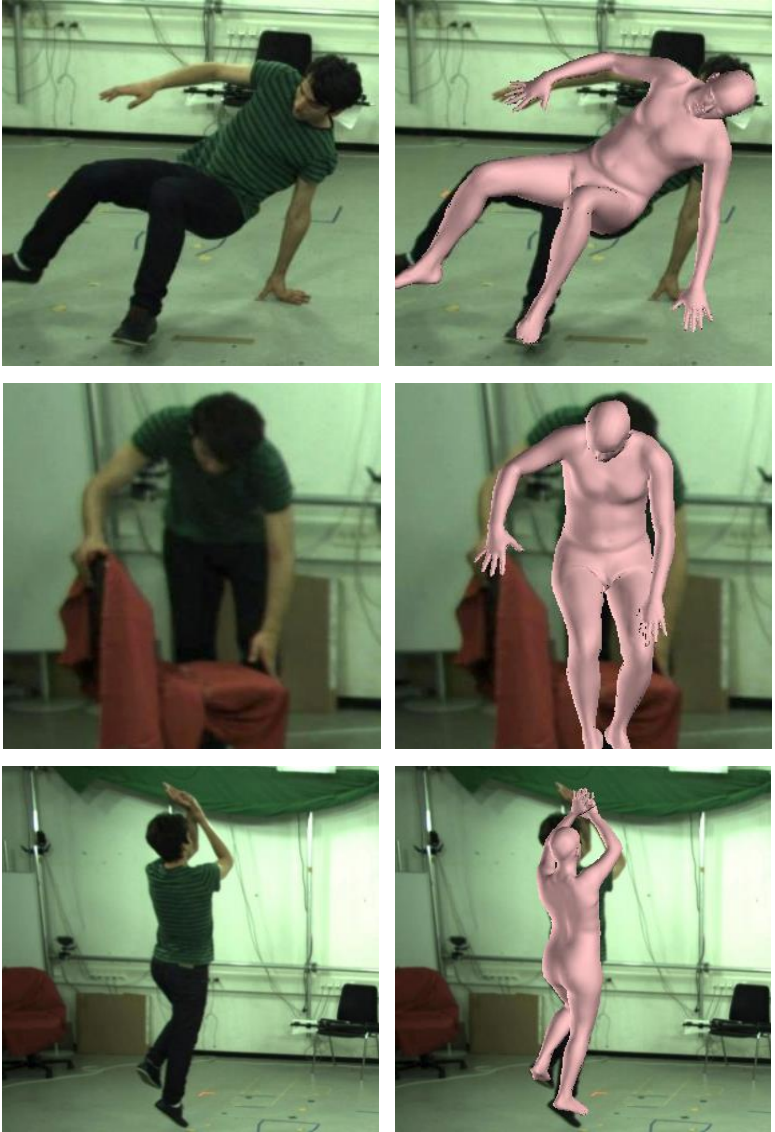} \\
   	(a) LSP & (b) Human3.6M  & (c) MPI-INF-3DHP \\
\end{tabular}
\caption{Qualitative results on challenging 2D images from the LSP~\cite{johnson2010clustered}, Human3.6M~\cite{h36m_pami}, and MPI-INF-3DHP~\cite{mehta2017monocular} datasets.
}
\label{Figure:qualitative_result}
\end{figure*}

Figure \ref{Figure:qualitative_result} includes qualitative results of our approach from different datasets involved in our evaluation, and we get excellent reconstruction even on some difficult gestures.

\subsection{Ablation Studies}
We evaluate the effectiveness of dictionary learning. 
We can extend the dictionary learning scheme readily to shape learning to achieve better results. 

To eliminate the influence of different training datasets, We follow the setting \cite{omran2018neural} and train our model only on the Human3.6M dataset. We do not use any shape annotations during network training. Table~\ref{tab:trained-on-h36m} shows that equipped with the proposed kinematic dictionary, our method even performs better than the methods trained with shape annotations.
Since we propose the shape dictionary together with the kinematic dictionary, we study their effectiveness in Table~\ref{tab:Dictionary}. We measure the surface error on UP-3D dataset. We can see kinematic dictionary is key to the performance. Adding shape dictionary can yield slightly better results.

In supplementary material, we also compare our choice of sparsemax with other options in dictionary learning. All in all, sparsemax could encourage the dictionary to cover the data better and improve the performance of pose estimation.

\begin{table}[t]
\small
\begin{center}
\scalebox{0.8}{
\setlength\tabcolsep{2mm}
\begin{tabular}{|l|c|c|}
\hline
Method &Training Data& P2$\downarrow$ \\
\hline
$\text{HMR}^1$~\cite{kanazawa2018end} &\tabincell{c}{  Human3.6M + MPI-INF-3DHP \\+ LSP + MPII + COCO}& 56.8 \\
$\text{HMR}^2$~\cite{kanazawa2018end} & Human3.6M & 77.6 \\
NBF  \cite{omran2018neural} & Human3.6M & 59.9\\
Ours(Semi) &Human3.6M&  \textbf{54.5}\\
\hline
\end{tabular}}
\end{center}
\caption{Influence of training data. Trained only on the Human3.6M dataset. Using the same dataset, our method is still the best, even better than the methods that use much more training data.}
\label{tab:trained-on-h36m}


\end{table}

\begin{table}[!t]
\small
\begin{center}
\scalebox{0.9}{
\setlength\tabcolsep{2mm}
\begin{tabular}{|l|c| c | c | c |}
\hline
Method & Shape D & Kinematic D & Surface Error$\downarrow$  & $\Delta$ \\
\hline
Baseline  & 	       	& 				& 141.2 & -\\ 
+ shape D & \checkmark   &  & 141.0  & -0.2\\
+ Kinematic D &  &  \checkmark  & 114.5 & -26.7\\
+ Both &\checkmark &\checkmark & \textbf{110.1} &  -31.1 \\
\hline
\end{tabular}}
\end{center}
\caption{Ablation results of different dictionaries. With both the kinematic and shape dictionaries, our method achieves the best result. }
\label{tab:Dictionary}
\vspace{-2mm}
\end{table}



\section{Conclusion}

In this paper, we propose a novel prior, the kinematic dictionary, to constrain the kinematic rotations of the human joints. We embed the kinematic dictionary into the deep learning framework. To make the dictionary learning consistent with the utilization of the dictionary in the network, we propose a novel objective function and introduce the sparsemax into the learning algorithm. Our method achieves competitive results against the state-of-the-art on relevant benchmarks. 

\bibliography{kinematic}

\end{document}


\linenumbers  %

\maketitle
This supplementary material provides additional details that were not included in the paper. We provide more information about the training data for 3D body reconstruction. 
We also show more qualitative results of our method and analyze failure cases as well. Then we give the direction of our future work.

\section{More Details}
\subsection{Training Details}
During training, we exploit the training datasets with both 2D and 3D keypoint annotations. We select images whose major keypoints are annotated and obtain training sets of 450k images for the 3D datasets and 55k images for the 2D datasets. Table~\ref{tab:dataset_size} shows the sizes of different datasets in training. Note that, although we use much less 2D data than HMR~\cite{kanazawa2018end}, which uses 110k images from in-the-wild datasets, we achieve better results thanks to our kinematic dictionary.

We use the regular train and test splits of these datasets and select images whose major keypoints are annotated.
We select images whose major keypoints are annotated, and set the sizes of the training sets to 300k, 150k, 7k for 3D datasets, and to 1k, 10k, 15k, and 30k for 2D datasets, respectively. 
Table \ref{tab:dataset_annotation} summarizes the training data as well as their annotations in details.


We use ground truth bounding boxes to crop subjects in all datasets as previous work did. All the cropped images are scaled to 224 $\times$ 224 and are randomly scaled, rotated, and flipped. In the 3D datasets, we use all the annotated keypoints. However, the definition of the joints in SMPL does not align perfectly with the joints defined in the datasets, so we follow Bogo \etal and Lassner \etal~\cite{bogo2016keep,lassner2017unite} to use a regression matrix of SMPL to get the 14 joints of LSP and incorporate five face keypoints from MS COCO following Kanazawa~\cite{kanazawa2018end}.

\begin{table}[t]
\begin{center}
\scalebox{0.8}{
\setlength\tabcolsep{5mm}
\begin{tabular}{|l|c| }
\hline
Dataset & Size \\
\hline
Human3.6M~\cite{h36m_pami}& 312k\\
UP-3D~\cite{lassner2017unite} & 7k \\
MPI-INF-3DHP~\cite{mono-3dhp2017}& 147k \\
LSP~\cite{Johnson11}& 1k \\
LSP-extended~\cite{johnson2010clustered}& 10k \\
MPII~\cite{andriluka20142d}& 14k \\
COCO~\cite{lin2014microsoft}& 28k \\
\hline
\end{tabular}
}
\end{center}
\vspace{-4mm}
\caption{Overview of training data sizes.  }
\label{tab:dataset_size}
\end{table}

\begin{table}[t]
\begin{center}
\begin{tabular}{|l|c c c|}
\hline
Dataset &SMPL &3D & 2D \\
\hline
Human3.6M&\checkmark &\checkmark &\checkmark \\
UP-3D & \checkmark	&  &\checkmark\\
MPI-INF-3DHP & &\checkmark &\checkmark\\
In-the-wild& & &\checkmark\\
\hline
\end{tabular}
\end{center}
\vspace{-4mm}
\caption{Training data overview. The in-the-wild dataset contains all the images from the LSP, LSP-extended, MPII and COCO dataset. }
\label{tab:dataset_annotation}
\end{table}

We initialize the encoder with pretrained weights on ImageNet \cite{deng2009imagenet}  and train the MLP from scratch. We use the Adam optimizer with the batch size of 128 and the learning rate of 1e-5. We first train the model on 3D datasets, and then on both 3D and in-the-wild 2D datasets. Each stage takes 200 epochs.

We transfer $\bm\Theta$ to the rotation matrix when computing the loss.
In our experiments, we use multiple datasets with different levels of annotations. Among the datasets, the 2D annotations are always present so that the loss term $L_{2D}$ is always used during training. The other loss terms $L_{3D}$ are added according to the availability of corresponding annotations.

\begin{figure*}[t]
\begin{center}
\includegraphics[width=0.9\textwidth]{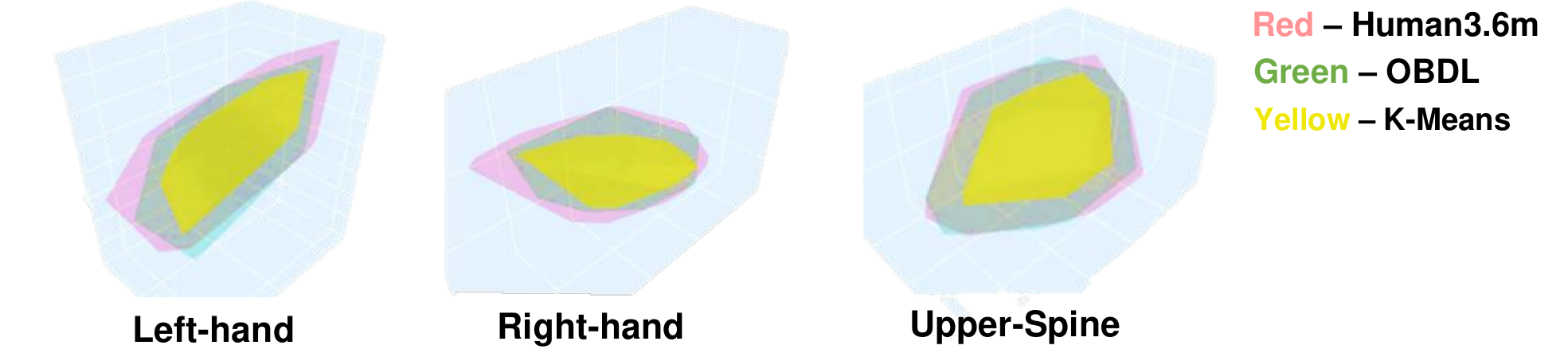}
\end{center}
\vspace{-5mm}
\caption{We plot in red the convex hull spanned by the rotation $\bm{\theta}$ of the Human3.6M dataset, in green the convex hull of our learned kinematic dictionary, and yellow the K-Means. Our OBDL method can better approximate the potential rotation space than K-Means does.}
\label{Figure:convex_hull}
\vspace{-5mm}
\end{figure*}

\section{Ablation Study on Dictionary Learning}

\subsection{Comparison of the Representation}
 We adopt a very different rotation representation, \ie, quaternions, to describe the SMPL pose parameters ${\bm\Theta}$. Through quaternions,  we are able to use Lerp to replace Slerp in the rotation space $SO(3)$.
Compared with other representations, such as axis-angle used by Kanazawa~\etal~\cite{kanazawa2018end}, rotation matrix used by Omran~\etal~\cite{omran2018neural} and 6D representation used by Kolotouros~\etal~\cite{kolotouros2019learning}, we show in Table~\ref{tab:Representation} that quaternions are the most suitable for dictionary learning. We use the same dictionary learning method described in the paper to learn the kinematic dictionaries for each representation respectively. We train the neural networks embedded with different kinematic dictionaries on the Human3.6M dataset, and report the results. Note that while kinematic dictionaries are learned from different representations, the shape dictionary used remains the same. We can see from Table~\ref{tab:Representation} that quaternions are key to the success of kinematic dictionary learning.

\begin{table}[t]
\begin{center}
\setlength\tabcolsep{2mm}
\begin{tabular}{|l| c | c |}
\hline
Representation    & Dimension & P2 $\downarrow$ \\
\hline
Quaternions(Ours) & 4D  & \textbf{53.4}  \\
Axis-angle       & 3D  & 83.0   \\
6D representation & 6D   & 112.1  \\
Rotation matrix   & 9D   & 192.4 \\
\hline
\end{tabular}
\end{center}
\vspace{-3mm}
\caption{Different representation of angle. We can see that quaternions perform favorably against other representations to a great extent.}
\label{tab:Representation}
\end{table}

\subsection{Comparison with Clustering}

\begin{table}[t]
\centering
\setlength\tabcolsep{2mm}
\begin{tabular}{|l | c | c|}
\hline
Method    & Dictionary Size & Ratio of Coverage\\
\hline
K-means & 128  & 0.89 \\
OBDL & 128 &\textbf{0.96}  \\
\hline
\end{tabular}
\caption{Comparison of different dictionary learning schemes for rotation recovery on the Human3.6m dataset.}
\label{tab:OBDL}
\vspace{-5mm}
\end{table}

Clustering methods are usually considered as the baseline to compare with dictionary learning. So we compare our method with K-means both statistically in Table~\ref{tab:OBDL} and visually in Figure~\ref{Figure:convex_hull} to show that our OBDL can better approximate the potential rotation space of human with a larger coverage of rotations on the Human3.6M dataset.

\subsection{Design Options}
 We validate our design choices in dictionary learning. We adopt a very different rotation representation, quaternions, to describe the SMPL pose parameters ${\bm\Theta}$. By exploiting the merits of quaternions,  we alternatively use Lerp to replace Slerp in the rotation space $SO(3)$. 



In dictionary learning,  we introduce the sparsemax function into sparse coding in the consistency of the usage of dictionaries in the deep learning framework. We find that though softmax can approximate sparse coding by enforcing most of the output elements to be close to dictionary atoms, using sparsemax to clip the small outputs can further reduce the error of approximation.

Table~\ref{tab:Function-dic} shows the superiority of sparsemax over softmax. In the table, we also show that the kinematic dictionary learned by OBDL which approximates the space of the potential rotation better than K-means directly influences the reconstruction results. Overall, using OBDL to learn dictionaries helps to achieve better results.

\begin{table}[t]
\begin{center}
\setlength\tabcolsep{2mm}
\begin{tabular}{|l|c  c |c |}
\hline
Exp & Function  & Dictionary Learning  & P2 $\downarrow$ \\
\hline
Ours & Sparsemax & OBDL &  \textbf{48.6} \\
Soft & Softmax  &  OBDL&  54.9\\ 
Cluster & Sparsemax & K-means & 53.4 \\ 
\hline
\end{tabular}
\end{center}
\vspace{-3mm}

\caption{Softmax vs sparsemax and K-means vs OBDL. With the sparsemax function and OBDL, our method achieves the best result on Human3.6M. }
\vspace{-5mm}

\label{tab:Function-dic}
\end{table}

\section{More Qualitative Results}

Here we want to show the reconstructed meshes, the estimated 3D keypoints, and the part segmentation in Figure~\ref{Figure:qualitative_result}, which can be directly obtained from the meshes, for in-the-wild images. We paste the recovered human back to the original uncropped images. We also show a video using our methods on the 3DPW dataset, which is an in-the-wild dataset that we neither use to learn dictionaries nor train neural networks. By transferring our dictionary to this {\it unseen} dataset, we show that our dictionaries learn the general human kinematic and shape knowledge and are thus transferable.

\begin{figure*}[h]
\begin{center}
\includegraphics[width=\textwidth]{lsp quality results_supp_1.pdf}
\end{center}
\caption{Qualitative results on challenging 2D images from the LSP dataset~\cite{johnson2010clustered}.}
\label{Figure:qualitative_result}
\end{figure*}


\begin{figure*}[t]
\begin{center}
	\includegraphics[width=0.9\textwidth]{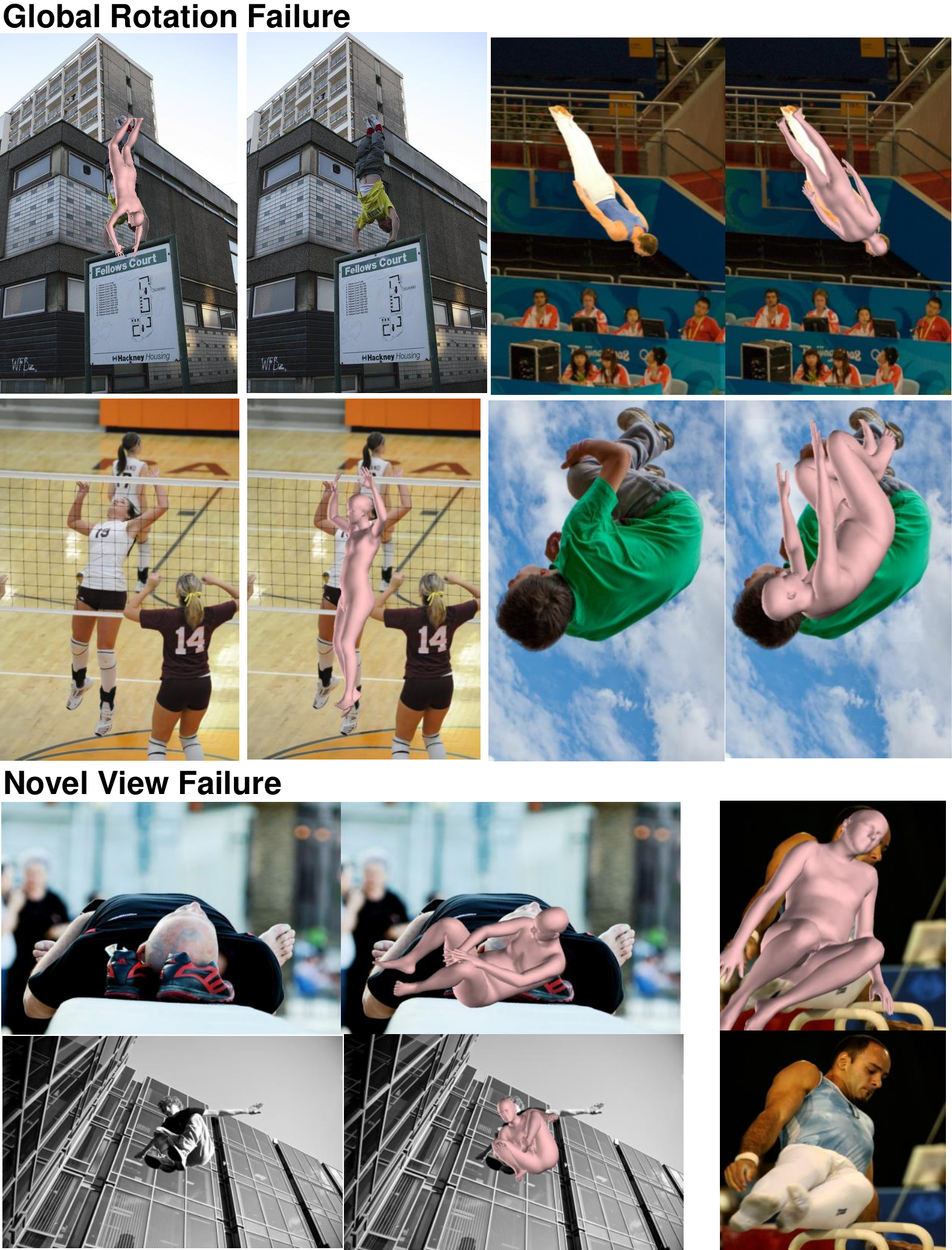}
\end{center}
\vspace{-5mm}
\caption{Failure cases. We can see inaccurate predictions of camera parameters make all keypoints shifted. And from novel view the reconstruction fails because the shortage of such training data.  }
\label{Figure:fail_cases}
\vspace{-5mm}
\end{figure*}

\section{Failure Cases}
\label{sec:failure}

Here we also show some negative results predicted by the network in Figure~\ref{Figure:fail_cases}. We can see the most significant errors are due to failures in the camera parameter regression. When the global rotation of the human mesh is inaccurate, all the keypoints would shift, resulting in large errors. This shows the importance of improving camera regression in future work. Another failure case is due to the uncommon viewing angles that rarely occur in the training data. 

\section{Future Work}

Our work tries to integrate sparse coding with deep learning, which could lead to many interesting works in the future. We will further explore if dictionaries can be online-updated when training neural networks. 
Furthermore, as stated in Section~\ref{sec:failure}, we will also seek for better camera regression methods to prevent negative results.

\clearpage

\bibliography{egbib}